# Maximum Likelihood Directed Enumeration Method in Piecewise-Regular Object Recognition

Andrey Savchenko

**Abstract**

We explore the problems of classification of composite object (images, speech signals) with low number of models per class. We study the question of improving recognition performance for medium-sized database (thousands of classes). The key issue of fast approximate nearest-neighbor methods widely applied in this task is their heuristic nature. It is possible to strongly prove their efficiency by using the theory of algorithms only for simple similarity measures and artificially generated tasks. On the contrary, in this paper we propose an alternative, statistically optimal greedy algorithm. At each step of this algorithm joint density (likelihood) of distances to previously checked models is estimated for each class. The next model to check is selected from the class with the maximal likelihood. The latter is estimated based on the asymptotic properties of the Kullback-Leibler information discrimination and mathematical model of piecewise-regular object with distribution of each regular segment of exponential type. Experimental results in face recognition for FERET dataset prove that the proposed method is much more effective than not only brute force and the baseline (directed enumeration method) but also approximate nearest neighbor methods from FLANN and NonMetricSpaceLib libraries (randomized kd-tree, composite index, perm-sort).

## 1. Introduction

Conventional machine learning techniques (support vector machines, multilayered feed-forward neural networks, deep neural networks, etc) [1] require large representative training sample to estimate the class border. These methods are known to be characterized with low accuracy if only few models are available for each class [2]. This issue is quite acute in, e.g., face recognition task in which it is sometimes difficult to gather various photos of the interesting person [2, 3]. The problem of insufficient accuracy becomes more complicated if the number of classes is large (hundreds or even thousands of classes). As a result, there is practically no alternative to the nearest neighbor (NN) methods in this task [1]. However, if the complex objects should be recognized in real-time (e.g., video-based face recognition [3]) and only standard hardware is available, the performance of brute-force implementation of the NN search is not enough. It seems that conventional fast approximate NN methods for image recognition, e.g. triangle tree [4], composite kd-tree [5], randomized kd-tree [6], Best-Bin First [7], etc. can be applied. Unfortunately, it is known that these techniques show good performance only if the first NN is quite different from other models [7]. Such restriction has much in common with many real-world applications, for instance, faces have similar shape and common features. Their other limitation is the application with similarity measures which satisfy metric properties (sometimes, triangle inequality and, usually, symmetry) [4, 7, 8].

Moreover, these methods are usually developed to approximately match very-large number (100 000) of image descriptors of extracted keypoints [9]. Hence, their performance is comparable with brute-force method for medium-sized vocabularies (thousands of classes). To decrease the recognition speed for such training sets, ordering permutations (perm-sort) method has recently been proposed [10]. Another interesting approach, namely, the directed enumeration method (DEM) outperforms the known approximate NN methods in face recognition [11].

Final issue is the heuristic nature of most popular approximate NN methods. It is practically impossible to prove that particular algorithm is optimal (in some sense) and nothing can be done to improve it. In this paper we propose an alternative solution on the basis of the statistical approach - while looking for the NN for particular query object, conventional probability of belonging of previously checked models to each class is estimated. The next model from the database is selected from the class with maximal probability. Thus, our task is to estimate this probability and to clarify the mentioned greedy-search algorithm.

The rest of the paper is organized as follows. In Section 2 we explore the task of recognition of piecewise-regular objects and present the statistical parametric criterion based on the Kullback-Leibler minimum discrimination principle [12]. In Section 3 we briefly review the baseline method (DEM), remind the asymptotic properties of the Kullback-Leibler discrimination and propose the novel Maximum-Likelihood DEM (ML-DEM). In Section 4 we demonstrate experimental results of comparison of our method with several approximate NN algorithms in face recognition with FERET dataset. Finally, concluding comments are given in Section 5.

## 2. Statistical recognition of piecewise-regular object

In the classification task it is required to assign the query object $X$ (facial photo, speech signal, image of natural scenes, text) to one of $R>1$ classes. Most part of contemporary research assumes that each class is specified by the given database $\{X_r\}$, $r \in \{1,...,R\}$ of $R$ cases (models).

Let the query object $X$ be represented as a sequence of $K$ regular (homogeneous) parts [13] extracted by any segmentation procedure: $X = \{X(k) | k = \overline{1,K}\}$. Every $k$-th segment $X(k) = \{\mathbf{x}_j(k) | j = \overline{1,n(k)}\}$ is put in correspondence with a sequence of (primitive) feature vectors $\mathbf{x}_j(k) = \{x_{j;1}(k),...,x_{j;p}(k)\}$ with fixed dimension $p=const$, where $n(k)$ is the number of features in the $k$-th segment. Similarly, every $r$-th model is represented as a sequence $X_r = \{X_r(k) | k = \overline{1,K_r}\}$ of $K_r$ segments and the $k$-th segment is defined as $X_r(k) = \{\mathbf{x}_j^{(r)}(k) | j = \overline{1,n_r(k)}\}$ of feature vectors $\mathbf{x}_j^{(r)}(k)$. Here $n_r(k)$ is the number of features in the $k$-th segment of the $r$-th model.

To apply statistical approach, let's assume that:

1. Vectors $\mathbf{x}_j(k)$, $\mathbf{x}_j^{(r)}(k_1)$ are *random*.

2. Segments $X(k), k=\overline{1,K}$ and $X_r(k), k=\overline{1,K_r}$ are *groups* - random samples of i.i.d. feature vectors $\mathbf{x}_j(k)$ and $\mathbf{x}_j^{(r)}(k)$, respectively.

3. Feature vectors of particular segment of one class are identically distributed.

As the procedure of automatic segmentation is inaccurate, every segment $X(k)$ should be compared with a set $N_r(k)$ of numbers of closed to $k$ segments of the $r$-th model. This neighborhood is determined for a specific task individually. If it is assumed that segmentation procedure is always correct, we may put $N_r(k) = \begin{cases} \{k\}, & K = K_r \\ \varnothing, & K \neq K_r \end{cases}$.

There are two possible approaches to estimate unknown class densities, namely, parametric and nonparametric [1]. Let's discover *parametric* approach in detail. It is assumed that distributions of vectors $\mathbf{x}_j(k)$ and $\mathbf{x}_j^{(r)}(k)$ are of multivariate *exponential type* $f_{\theta;n}$ [12] generated by the fixed (for all classes) function $f_0(\cdot)$ with $p$-dimensional parameter vector $\mathbf{\theta}$:

$$f_{\theta;n}(\widetilde{X}) = \exp(\mathbf{\tau(\theta)} \cdot \hat{\mathbf{\theta}}(\widetilde{X})) \cdot f_0(\widetilde{X}) / M(\mathbf{\tau}) \tag{1}$$

where $\hat{\mathbf{\theta}}(\widetilde{X})$ is an estimation of parameter $\mathbf{\theta}$ using available data (random sample) $\widetilde{X}$ of size $n$,

$$M(\mathbf{\tau}) = \int \exp(\mathbf{\tau(\theta)} \cdot \hat{\mathbf{\theta}}(\widetilde{X})) \cdot f_0(\widetilde{X}) d\widetilde{X} \tag{2}$$

and $\mathbf{\tau(\theta)}$ is a normalizing function ($p$-dimensional parameter vector) defined by the following equation if the parameter estimation $\hat{\mathbf{\theta}}(\widetilde{X})$ is unbiased (see [12] for details)

$$\int \hat{\mathbf{\theta}}(\widetilde{X}) \cdot f_{\theta;n}(\widetilde{X}) d\widetilde{X} \equiv \frac{d}{d\mathbf{\tau}} \ln M(\mathbf{\tau}) = \mathbf{0} \tag{3}$$

Each $r$-th class of each $k$-th segment is determined by parameter vector $\mathbf{\theta}_r(k)$. This assumption about exponential family $f_{\hat{\theta}(X_r(k));n(k)}$ in which parameter $\mathbf{\theta}_r(k)$ is estimated by using the observed (given) sample $X_r(k)$, covers wide range of known distributions (polynomial, normal, etc.) [12].

Hence, the recognition task is reduced to a problem of statistical testing of $R$ *simple* hypothesis about parameter vector $\mathbf{\theta}_r(k)$. In this paper we focus on the case of full prior uncertainty and assume that the prior probabilities of each class are equal. In such case, Bayesian approach will be equivalent to the maximum likelihood criterion. For our task, every segment is recognized with the following rule

$$\max_{k_1 \in N_r(k)} f_{\hat{\theta}(X_r(k_1));n(k)}(X(k)) \to \max_{r \in \{1,\ldots,R\}}. \tag{4}$$

It can be shown that eq. (4) is equivalent to the Kullback-Leibler minimum information discrimination principle [12]

$$\rho(X, X_r) \to \min_{r = 1, R}, \tag{5}$$

where

$$\rho(X, X_r) = \\ = \frac{1}{nK} \sum_{k=1}^{K} \min_{k_1 \in N_r(k)} \hat{I}\left(*: f_{\hat{\theta}(X_r(k_1));n(k)}; X(k)\right) \tag{6}$$

and

$$\hat{I}\left(*: f_{\hat{\theta}(X_r(k_1)), n(k)}; X(k)\right) = $$
$$= \int f_{\hat{\theta}(X(k)), n(k)}(\widetilde{X}) \cdot \ln \frac{f_{\hat{\theta}(X(k)), n(k)}(\widetilde{X})}{f_{\hat{\theta}(X_r(k_1)), n(k)}(\widetilde{X})} d\widetilde{X}$$
(7)

is the Kullback-Leibler divergence between segments $X(k)$ and $X_r(k_1)$; and $n = \sum_{k=1}^{K} n(k)$.

Thus, criterion (5)-(7) is an obvious implementation of Bayesian approach to composite object recognition if the probabilistic mathematical model of piecewise-regular object [13] is used.

### 3. Maximum-likelihood directed enumeration method

Let's use an approach known from artificial intelligence to create an approximate NN algorithm for measure of similarity (6), (7). Namely, we primarily focus on *greedy* algorithms: on each step it explores the model which is the NN of the query object $X$ with the highest probability. Such choose of the greedy class of algorithms is explained not only by its simplicity, but by the fact that practically all known approximate NN methods are greedy in some sense.

#### 3.1. Baseline: directed enumeration method

As a baseline method we use the DEM [11] which was based on the metric properties of the Kullback-Leibler divergence and regards the models' similarity $\rho_{i,j} = \rho(X_i, X_j)$ as an average information from an observation to distinct class *i* from an alternative class *j*. Hence, at the preliminarily step of the DEM, the model distance matrix $P = [\rho_{i,j}]$ is calculated as it is done in the AESA (Approximating and Eliminating Search Algorithm) method [4]. This time-consuming procedure should be repeated only once for a particular task and training set.

Original DEM used the following heuristic: if there exists a model $X^*$ for which $\rho(X, X^*) < \rho_0 \ll 1$, then for an arbitrary *r*-th model the following condition holds $|\rho(X, X_r) - \rho(X^*, X_r)| \ll 1$ with high probability. Hence, the criteria (5) can be simplified

$$\rho(X, X^*) < \rho_0 = const.$$
(8)

Eq. (8) defines the termination condition of the approximate NN method. If false-accept rate (FAR) is fixed $\beta = const$, then $\rho_0$ is evaluated as a $\beta$-quantile of the distances between images from distinct classes $\{\rho_{i,j} | i = \overline{1,R}, j = \overline{1,R}, i \neq j\}$. As a matter of fact, the optimization task (5) is replaced to an exhaustive search which terminates if condition (8) holds for the currently checked model.

According to the DEM [11], at first, the model $X_{r_1}, r_1 \in \{1,...,R\}$ is randomly chosen so that

$$\forall i, j \in \{1,...,R\} \rho_{i,r_1} \neq \rho_{j,r_1},$$
(9)

and the distance $\rho\left(X, X_{r_1}\right)$ is calculated. If the distance is lower than a threshold $\rho_0$ (8), the search is terminated. Otherwise, it is put into the priority queue of models sorted by the distance to $X$. Next, the highest priority item $X_i$ is pulled from the queue and the set of models $X_i^{(M)}$ is determined from

$$\left(\forall X_j \notin X_i^{(M)}\right)\left(\forall X_k \in X_i^{(M)}\right), \quad \Delta\rho(X_j) \geq \Delta\rho(X_k) \tag{10}$$

where $\Delta\rho(X_j) = |\rho_{i,j} - \rho(X, X_j)|$ is the deviation of $\rho_{i,j}$ relative to the distance between $X$ and $X_j$. For all models from the set $X_i^{(M)}$ the distance to the query object is calculated and the condition (9) is verified. After that, every previously unchecked model from this set is put into the priority queue. The method is terminated if for one model object condition (9) holds or after checking for $E_{\max} = const$ models.

As we stated earlier, this method is heuristic as most popular approximate NN algorithms. However, the probability that the model is the NN of $X$ can be directly calculated for the Kullback-Leibler discrimination by using its asymptotic properties. Let's describe them briefly.

### 3.2. Asymptotic properties

It is known [12] that if the segment $X(k)$ has distribution of exponential type with parameter $\hat{\boldsymbol{\theta}}(X_v(k)), v \in \{1,...,R\}$, then the 2-times Kullback-Leibler divergence (6) $2\hat{I}\left(*: f_{\hat{\boldsymbol{\theta}}(X_r(k_1)), n(k)}; X(k)\right)$ is asymptotically distributed as a noncentral $\chi^2$ with $p$ degrees of freedom and noncentrality parameter $2\hat{I}\left(*: f_{\hat{\boldsymbol{\theta}}(X_r(k_1)), n(k)}; X_v(k)\right)$. By assuming the independence of all $K$ segments $X(k)$, we can conclude that if the query object $X$ corresponds to class $v$, then the distance $2nK \cdot \rho(X, X_v)$ is asymptotically distributed as a $\chi^2$ with $K \cdot p$ degrees of freedom. Similarly, $2nK \cdot \rho(X, X_r)$, $r \neq v$ has asymptotic non-central $\chi^2$ distribution with $K \cdot p$ degrees of freedom and noncentrality parameter $2nK \cdot \rho_{v,r}$. If $K \cdot p$ is high, then, by using the central limit theorem we obtain the normal distribution

$$N\left(\rho_{v,r} + \frac{p}{2n}; \frac{\sqrt{8nK \cdot \rho_{v,r} + 2K \cdot p}}{2nK}\right). \tag{11}$$

of the distance $\rho(X, X_r)$.

### 3.3. Proposed method

Based on the asymptotic distribution (11) we replace the step (10) of the original DEM to the procedure of choosing the maximum likelihood model. Let's assume that the models $X_{r_1}, ..., X_{r_l}$ have been checked before the $l$-th step, i.e. the distances $\rho(X, X_{r_1}), ..., \rho(X, X_{r_l})$ have been calculated. By assuming the equal prior probability of each class and independence of the models from different classes, let's choose the next most probable model $X_{r_{l+1}}$ with the maximum likelihood method [1]:

$$r_{l+1} = \underset{v \in \{1,...,R\} - \{r_1,...,r_l\}}{\arg\max} \prod_{i=1}^{l} f\left(\rho(X, X_{r_i}) | W_v\right), \tag{12}$$

where $f\left(\rho(X, X_{r_i}) | W_v\right)$ is the conditional density (likelihood) of the distance $\rho(X, X_{r_i})$ if the hypothesis $W_v$ is true (the class label of the query object $X$ is $v$). To estimate this likelihood, asymptotic distribution (11) is used. Hence, the likelihood in (12) can be written in the following form

$$f\left(\rho(X, X_{r_i}) | W_v\right) = \frac{2nK}{\sqrt{2\pi \cdot (8nK \cdot \rho_{v,r_i} + 2K \cdot p)}} \times \exp\left[-\frac{(2nK \cdot (\rho(X, X_{r_i}) - \rho_{v,r_i}) - K \cdot p)^2}{8nK \cdot \rho_{v,r_i} + 2K \cdot p}\right] =$$
$$= \frac{2nK}{\sqrt{2\pi}} \exp\left[-\frac{1}{2} \ln(8nK \cdot \rho_{v,r_i} + 2K \cdot p)\right] \times \exp\left[-\frac{(2nK \cdot (\rho(X, X_{r_i}) - \rho_{v,r_i}) - K \cdot p)^2}{8nK \cdot \rho_{v,r_i} + 2K \cdot p}\right] \tag{13}$$

By dividing (12) by a constant $\left(2nK / \sqrt{2\pi}\right)^l$, taking a natural logarithm, dividing by $nK/2$ and adding $l$, expression (12) can be finally transformed to

$$r_{l+1} = \underset{\mu \in \{1,...,R\} - \{r_1,...,r_l\}}{\arg\min} \sum_{i=1}^{l} \varphi_\mu(r_i). \tag{14}$$

where

$$\varphi_\mu(r_i) = \frac{\left(\rho(X, X_{r_i}) - \rho_{\mu,r_i} - \frac{p}{2n}\right)^2}{\left(4\rho_{\mu,r_i} + \frac{p}{n}\right)} + \frac{1}{4nK} \ln\left(4\rho_{\mu,r_i} + \frac{p}{n}\right). \tag{15}$$

As the average segment's size is usually much higher the number of parameters $n \gg N$, then the function in (15) can be simplified

$$\varphi_\mu(r_i) \approx \frac{(\rho(X, X_{r_i}) - \rho_{\mu,r_i})^2}{4\rho_{\mu,r_i}} \tag{16}$$

This equation is in good agreement with the heuristic from the original DEM [11] - the closer are the distances $\rho(X, X_{r_i})$ and $\rho_{\mu,r_i}$ and the higher is the distance between models $X_\mu$ and $X_{r_i}$, the lower is $\varphi_\mu(r_i)$.

Next, the termination condition (8) is checked for the model $X_{r_{l+1}}$. If the distance $\rho(X, X_{r_{l+1}})$ is lower than a threshold $\rho_0$, then the search procedure is stopped on the $L_{checks} = l+1$ step. Otherwise the model $X_{r_{l+1}}$ is put into the set of previously checked models and the procedure (14), (16) is repeated.

Let's return to the initialization of our method. We would like to choose the first model $X_{r_1}$ to obtain the decision (8) in a shortest (in terms of number of calculations $L_{checks}$) way. Let's maximize an average probability to obtain the decision on the second step

$$r_1 = \arg\max_{\mu \in \{1,...,R\}} \frac{1}{R} \sum_{v=1}^{R} P\left(\varphi_v(\mu) \leq \min_{r \in \{1,...R\}} \varphi_r(\mu) \Big| W_v \right). \tag{17}$$

To estimate the conditional probability $P\left(\varphi_v(r_1) \leq \min_{r \in \{1,...R\}} \varphi_r(r_1) \Big| W_v \right)$ in (17) we use again the asymptotic distribution (11):

$$P\left(\varphi_v(\mu) \leq \min_{r \in \{1,...R\}} \varphi_r(\mu) \Big| W_v \right) = \prod_{r=1}^{R} P\left( \frac{(\rho(X,X_\mu) - \rho_{v,\mu})^2}{\rho_{v,\mu}} \leq \frac{(\rho(X,X_\mu) - \rho_{r,\mu})^2}{\rho_{r,\mu}} \Big| W_v \right)$$
$$= \prod_{r \in \{1,...,R | \rho_{r,\mu} \geq \rho_{v,\mu}\}} P\left(\rho(X,X_\mu) \leq \sqrt{\rho_{r,\mu} \cdot \rho_{v,\mu}} \Big| W_v \right) \times \prod_{r \in \{1,...,R | \rho_{r,\mu} < \rho_{v,\mu}\}} P\left(\rho(X,X_\mu) \geq \sqrt{\rho_{r,\mu} \cdot \rho_{v,\mu}} \Big| W_v \right) \tag{19}$$

Finally, based on (11) one can write

$$P\left(\varphi_v(\mu) \leq \min_{r \in \{1,...R\}} \varphi_r(\mu) \Big| W_v \right) = \prod_{r=1}^{R} \left( \frac{1}{2} + \Phi\left( \frac{\sqrt{nK}}{2} \left| \sqrt{\rho_{r,\mu}} - \sqrt{\rho_{v,\mu}} \right| \right) \right), \tag{20}$$

where $\Phi(\cdot)$ is the cumulative density function of the normal distribution. As a result, the first model to check $X_{r_1}$ is obtained from the following expression

$$r_1 = \arg\max_{\mu \in \{1,...,R\}} \sum_{v=1}^{R} \prod_{r=1}^{R} \left( \frac{1}{2} + \Phi\left( \frac{\sqrt{nK}}{2} \left| \sqrt{\rho_{r,\mu}} - \sqrt{\rho_{v,\mu}} \right| \right) \right). \tag{21}$$

Thus, the proposed ML-DEM (9), (14), (16), (21) is an optimal (maximal likelihood) greedy algorithm for an approximate NN search with termination condition (8) for the Kullback-Leibler discrimination (6), (7). As a matter of fact, this method can be applied with an arbitrary complex similarity measure. The next section provides experimental evidence to support this claim.

## 4. Experimental study

In this section we present experimental study of proposed ML-DEM in the problem of face recognition.

### 4.1. Face recognition

It is required to assign a query image $X$ to one of $R$ classes (persons) specified by reference images $\{X_r\}$, $r \in \{1,...,R\}$.

We assume that the object of interest (face) is preliminarily detected by an arbitrary algorithm (e.g., Viola-Jones method [14]). To implement here statistical approach described in Section 2, the image is divided into a regular grid of $S_1 \times S_2$ blocks, $S_1$ rows and $S_2$ columns (in our notation, $S_1 \cdot S_2 = K = K_1 = ... = K_R$). Next, the histogram $H^{(r)}(s_1,s_2) = [h_1^{(r)}(s_1,s_2),...,h_N^{(r)}(s_1,s_2)]$ of appropriate simple features is separately evaluated for each block $(s_1,s_2)$ of the reference image $X_r$. Here $N$ is the number of bins in the histogram, $s_1 \in \{1,...,S_1\}$, $s_2 \in \{1,...,S_2\}$. The most popular image point's simple feature is the gradient orientation (probably, weighted with the gradient magnitude), i.e., $H^{(r)}(s_1,s_2)$ is the histogram of oriented gradients (HOG) [15]. In this paper we assume, that each histogram $H^{(r)}(s_1,s_2)$ is normalized, so that it may be treated as a probability distribution [15]. The united vector $[H^{(r)}(1,1),...,H^{(r)}(1,S_2),H^{(r)}(2,1),...,H^{(r)}(S_1,S_2)]$ is amounted the desired descriptor of the whole reference image. The same procedure is repeated to evaluate the histograms $H(s_1,s_2) = [h_1(s_1,s_2),...,h_N(s_1,s_2)]$ corresponding to the query image.

The neighborhood ($N_r(k)$) of block $(s_1,s_2)$ contains the cells $(\tilde{s}_1,\tilde{s}_2)$ for which $|\tilde{s}_1 - s_1| < \Delta$, $|\tilde{s}_2 - s_2| < \Delta$, where $\Delta = const$ is chosen based on the concrete task (usually $\Delta$=0 or $\Delta$=1 [16]). In such case, the distance in the nearest neighbor rule (5), (6) will be calculated with the mutual alignment of the histograms in the $\Delta$- neighborhood as follows

$$\rho(X,X_r) = \frac{1}{S_1 S_2} \sum_{s_1=1}^{S_1} \sum_{s_2=1}^{S_2} \min_{\substack{|\Delta_1| \le \Delta, \\ |\Delta_2| \le \Delta}} \rho^{(H)}\left(H(s_1,s_2), H^{(r)}(s_1+\Delta_1, s_2+\Delta_2)\right), \quad (22)$$

where $\rho^{(H)}\left(H(s_1,s_2), H^{(r)}(s_1+\Delta_1, s_2+\Delta_2)\right)$ is an appropriate distance between HOGs. If the Kullback-Leibler discrimination (7) is applied, the distance between HOGs can be written in the following form

$$\rho_{KL}\left(H, H^{(r)}\right) = \sum_{i=1}^{N} h_{K;i} \ln \frac{h_{K;i}}{h_{K;i}^{(r)}}. \quad (23)$$

Here we missed indices $(s_1,s_2)$ for simplicity and use the convolution of the HOGs with any kernel $K_{ij}$

$$h_{K;i}^{(r)} = \sum_{j=1}^{N} K_{ij} h_j^{(r)}, \quad h_{K;i} = \sum_{j=1}^{N} K_{ij} h_j \quad (24)$$

to prevent division by zero in (23) if the histogram value for several bins is equal to zero. In the experiment we use the conventional Gaussian Parzen window [17].

In this study we also explore the homogeneity-testing probabilistic neural network (HT-PNN) showed good accuracy with HOGs in face recognition and known to be equivalent to the statistical approach if the pattern recognition problem is referred as a task of testing for homogeneity of segments [18]:

$$\rho_{HT-PNN}\left(H, H^{(r)}\right) = \sum_{i=1}^{N} \left( h_i \ln \frac{2 h_{K;i}}{h_{K;i} + h_{K;i}^{(r)}} + h_i^{(r)} \ln \frac{2 h_{K;i}^{(r)}}{h_{K;i} + h_{K;i}^{(r)}} \right). \quad (25)$$

### 4.2. Experimental results

In this experiment FERET dataset was used (http://www.itl.nist.gov/iad/humanid/feret/feret_master.html). *R=1432* frontal images of 994 persons populate the database (i.e. a training set), other 1288 frontal photos of the same persons formed a test set.

The faces were detected with the OpenCV library. The median filter with window size (3x3) was applied to remove noise in detected faces. The faces were divided into 100 fragments ($S_1 = S_2 = 10$). The number of bins in the HOG *N*=8. To obtain threshold $\rho_0$, the FAR is fixed to be $\beta = 1\%$. These parameters provide the best accuracy in our experiments.

The error rate obtained by cross-validation with the NN rule and similarity measure (1) with Euclidean and the PNNH (2) distances is shown in Table 1 in the format average error rate ± its standard deviation.

Table 1. Error rate (in %) of the NN method (22)

|  | $\Delta=0$ | $\Delta=1$ |
|---|---|---|
| **Kullback-Leibler (23)** | 8.9±1.3 | 7.0±1.3 |
| **HT-PNN (25)** | 7.8±1.2 | 6.6±1.3 |

From this table one could notice that, first, alignment of HOGs (22) with $\Delta = 1$ improves the recognition accuracy. And, second, we experimentally supports the fact [18] that the error rate for the Kullback-Leibler distance (23) exceeds the error for the HT-PNN (25).

In the next experiment we compare the performance of the proposed ML-DEM with an original DEM [11], brute force and several approximate NN methods from FLANN [5] and NonMetricSpaceLib [19] libraries showed the best speed, namely

1. Randomized kd-tree from FLANN with 4 trees [6]

2. Composite index from FLANN which combines the randomized kd-trees (with 4 trees) and the hierarchical k-means tree [5].

3. Ordering permutations (perm-sort) from NonMetricSpaceLib which is known to decrease the recognition speed for medium-sized databases (thousands of models) [10].

We evaluate the error rate (in %) and the average time (in ms) to recognize one test image with a modern laptop (4 core i7, 6 Gb RAM) and Visual C++ 2013 Express compiler (x64 environment) and optimization by speed. We explore an obvious way to improve performance by using parallel computing [16]. Namely, the whole training set was divided into *T*=const non-overlapped parts and each part is processed in its own task. All tasks work in parallel and terminate right after any task finds the solution. Each task is implemented as a separate thread by using the Windows ThreadPool API. We analyze both conventional nonparallel case (*T*=1) and the parallel one (*T*=8).

After several experiments the best (in terms of recognition speed) value of parameter *M* of original DEM (10) was chosen *M=64* for nonparallel case and *M=16* for parallel one. Parameter $E_{\max}$ was chosen to achieve the recognition accuracy which is not 0.5% less than the accuracy of brute force (Table 1). If such accuracy could not be achieved, $E_{\max}$ was set to be equal to the count of models assigned to each task.

The average recognition time per one test image (in ms) for the Kullback-Leibler discrimination (23) for Δ=0 and Δ=1 is shown in Fig. 1 and Fig. 2, respectively.

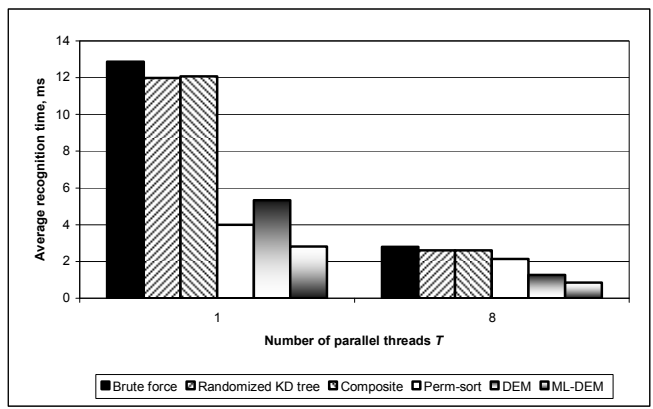

Figure 1: Average recognition time, Kullback-Leibler discrimination, Δ=0

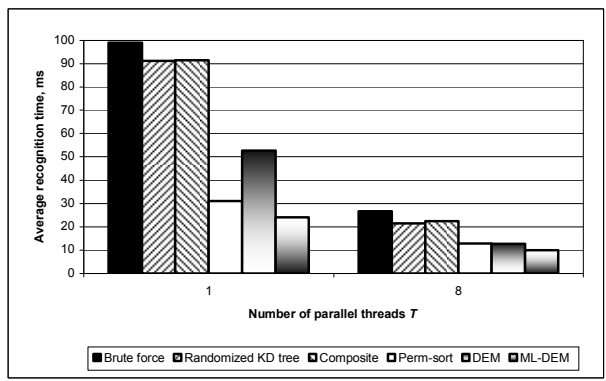

Figure 2: Average recognition time, Kullback-Leibler discrimination, Δ=1

Here one can notice that modifications of kd-tree from FLANN (randomized and composite indices) do not show superior performance even over brute force as the number of models in the database is not very high. However, as it was expected, perm-sort method is characterized with 2-3.5 times lower recognition speed in comparison with an exhaustive search. Moreover, perm-sort seems to be better than the original DEM for nonparallel case (*T*=1), though the DEM's parallel implementation is a bit better. The most important conclusion here is that the proposed ML-DEM shows the highest speed in all experiments.

To clarify the difference in performance of the original DEM and the proposed ML-DEM, we show the dependence of the error rate and the number of checked models $L_{checks}/R \cdot 100\%$ on the maximum number of models to be checked $E_{\max}$ in Fig. 3 and Fig. 4, respectively. We describe here the case Δ=1 for which the DEM is the best among all other methods.

Fig. 3 demonstrates that the speed of convergence to an optimal solution for the ML-DEM is much higher than the same indicator of the DEM. Even when $E_{\max} = 0.1 \cdot R$ we can get an appropriate solution.

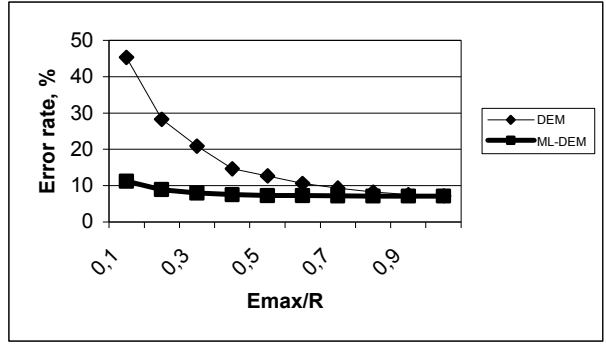

Figure 3: Dependence of error rate on $E_{\max}$, Kullback-Leibler discrimination, $\Delta=1$

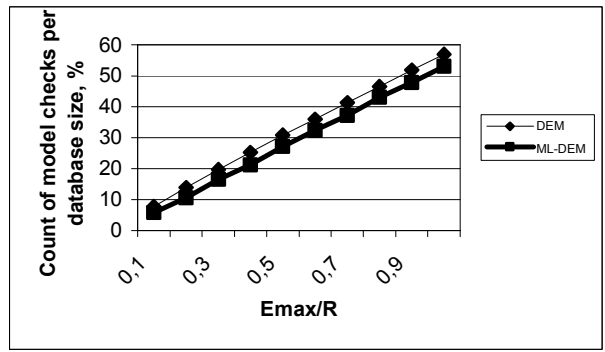

Figure 4: Count of models checks per database size $L_{checks} / R \cdot 100\%$, Kullback-Leibler discrimination, $\Delta=1$

Fig. 4 proves that the proposed ML-DEM is an optimal greedy algorithm in terms of the number of calculated distances $L_{checks}$. However, additional computations of the ML-DEM (14), (16) which include the calculations for every non previously checked model, are quite complex. Hence, the difference in performance with the DEM and other approximate NN methods is high only for very complex similarity measures (e.g., for the case of HOG's alignment, $\Delta=1$).

The average recognition time for the HT-PNN (25) for $\Delta=0$ and $\Delta=1$ is shown in Fig. 5 and Fig. 6, respectively.

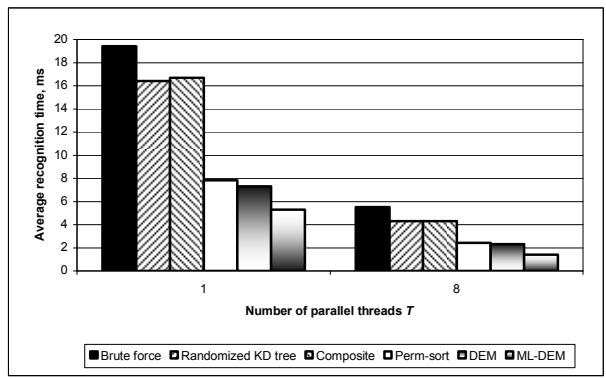

Figure 5: Average recognition time, HT-PNN, $\Delta=0$

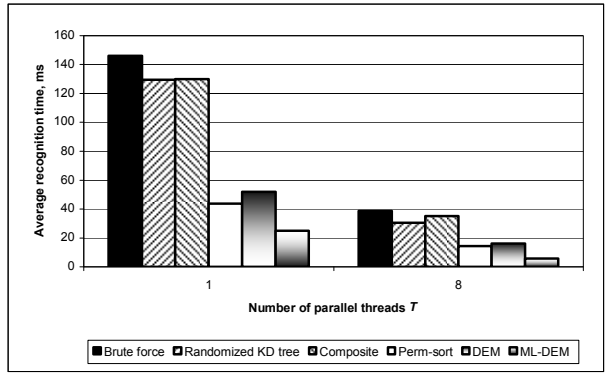

Figure 6: Average recognition time, HT-PNN, Δ=1

The results of this experiment are very similar to the Kullback-Leibler results (Fig. 1, 2) though the error rate here is 0.5-1% lower (see also Table 1). However, the original DEM is here a bit faster than the perm-sort for conventional distance (Δ=0, Fig. 5) but is not so effective for alignment (Δ=1, Fig. 6). FLANN's kd-trees are 10-15% faster than the brute force. And again, the proposed ML-DEM is the best choice here especially for most complex case ($T$=8, Δ=1) for which only 6 ms (in average) is necessary to recognize a query face with 93% accuracy.

## 5. Conclusion and future work

We have shown that using the asymptotic properties (11) of the Kullback-Leibler discrimination for recognition piecewise-regular objects (5)-(7) in the DEM [11] gives very good results in face recognition with medium-sized database, reducing the recognition speed by more than 2.5-6.5 times in comparison with brute force and by 1.2-2.5 times in comparison with other approximate NN methods from FLANN and NonMetricSpaceLib libraries. We studied the influence of various distance parameters (distance type, neighborhood size Δ) and the maximal number $E_{\max}$ of distances to calculate.

In contrast to the most popular fast algorithms, our method is not heuristic (except the termination condition (8)). Moreover, it does not build data structure based on an algorithmic properties of applied similarity measure (e.g., triangle inequality of Minkowski metric in the AESA [4], Bregman ball for Bregman divergences [8]). The proposed ML-DEM is an optimal (maximum likelihood) greedy method in terms of the number of distance calculations (see Fig. 3) for NN rule (5) with the sum of Kullback-Leibler discriminations (5), (6). Moreover, as we showed in the last part of our experimental study, the ML-DEM can be successfully applied (Fig. 5, 6) with other distances, e.g., not popular but very accurate HT-PNN (25) [18].

The main direction for further research in the ML-DEM can be related to improving the performance of each step (14), (16) by its simplification or the usage of ideas of pivot-based approximate NN methods [20]. As a matter of fact, it is the main obstacle to use our method with various similarity measures. We are also working on exploration the influence of the popular distances (Euclidean, chi-squared, Jensen-Shannon, etc.) on the performance of our method.

## References

[1] S. Theodoridis, and K. Koutroumbas (eds.). Pattern recognition. Boston: Academic Press, 2008. 840 p.

[2] X. Tan, S. Chen, Z. H. Zhou, and F. Zhang. Face recognition from a single image per person: a survey. Pattern Recognition, 39(9):1725–1745, 2006

[3] C. Shan. Face recognition and retrieval in video // Video Search and Mining, Studies in Computational Intelligence, 287:235-260, 2010

[4] E. Vidal. An algorithm for finding nearest neighbours in (approximately) constant average time. Pattern recognition Letters, 4(3):145–157, 1986.

[5] M. Muja, and D. G. Lowe. Fast approximate nearest neighbors with automatic algorithm configuration. 4th International Conference on Computer Vision Theory & Applications (VISAPP), 1: 331-340, 2009.

[6] C. Silpa-Anan, and R. Hartley. Optimised KD-trees for fast image descriptor matching. CVPR, Anchorage, Alaska, USA, pages 1-8, 2008

[7] J. Beis, and D. G. Lowe. Shape indexing using approximate nearest–neighbour search in high dimensional spaces. CVPR, San Juan, Puerto Rico, pages 1000–1006, 1997.

[8] L. Cayton. Efficient Bregman Range Search. Advances in Neural Information Processing Systems, Y. Bengio, D. Schuurmans, J.D. Lafferty, C.K.I. Williams, and A. Culotta (Eds.), 22:243-251, 2009

[9] D. Lowe. Distinctive image features from scale–invariant keypoints. International Journal of Computer Vision, 60(2):91–110, 2004

[10] E. C. Gonzalez, K. Figueroa, and G. Navarro. Effective Proximity Retrieval by Ordering Permutations. IEEE Transactions on Pattern Analysis and Machine Intelligence, 30(9): 1647–1658, 2008.

[11] A. V. Savchenko, Directed enumeration method in image recognition. Pattern Recognition, 45(8):2952–2961, 2012.

[12] S. Kullback. Information Theory and Statistics. Mineola, N.Y: Dover Publications, 1997.

[13] P. Prandoni, and M.Vetterli. Approximation and compression of piecewise smooth functions. Philosophical Transactions of the Royal Society, 357(1760):2573-2591, 1999.

[14] P. Viola, and M. Jones. Rapid object detection using a boosted cascade of simple features. CVPR, Kauai, HI, USA, pages 511–518, 2001.

[15] N. Dalal, and B. Triggs. Histograms of oriented gradients for human detection. CVPR, San Diego, CA, USA, pages 886–893, 2005.

[16] A. V. Savchenko. Real-Time Image Recognition with the Parallel Directed Enumeration Method. The 9th ICVS (International Conference on Vision Systems), M. Chen, B. Leibe, and B. Neumann (Eds.), LNCS, 7963:123-132, 2013.

[17] D. F. Specht. Probabilistic neural networks. Neural networks, 3(1):109–118, 1990.

[18] A. V. Savchenko. Probabilistic neural network with homogeneity testing in recognition of discrete patterns set. Neural Networks, 46:227–241, 2013.

[19] L. Boytsov, and N. Bilegsaikhan. Engineering Efficient and Effective Non-Metric Space Library. The 6th SISAP (Similarity Search and Applications), N. Brisaboa, O. Pedreira, P. Zezula (Eds.), LNCS 8199: 280-293, 2013

[20] B. Bustos, E. Ch´avez, and G. Navarro. Pivot selection techniques for proximity searching in metric spaces. Pattern Recognition Letters, 24-14: 2357–2366, 2003